\documentclass[runningheads]{llncs}
\usepackage[T1]{fontenc}
\usepackage{graphicx}
\usepackage{hyperref}
\usepackage{amsmath}
\usepackage{multirow}
\usepackage{fontawesome5}

\begin{document}
\title{Assessing LLMs Suitability for Knowledge Graph Completion}
%
%
\author{Vasile Ionut Remus Iga\orcidID{0009-0001-4568-929X} \and \\
Gheorghe Cosmin Silaghi\orcidID{0000-0002-3447-4736}\faIcon[regular]{envelope} }
\authorrunning{V.I.R. Iga \& G.C. Silaghi }
%
\institute{Babe\c{s}-Bolyai University \\ Business Informatics Research Center, Cluj-Napoca, Romania  \\	
\email{vasile.iga@ubbcluj.ro, \faIcon[regular]{envelope} gheorghe.silaghi@ubbcluj.ro}
}%
\maketitle              
\begin{abstract}
Recent work has shown the capability of Large Language Models (LLMs) to solve tasks related to Knowledge Graphs, such as Knowledge Graph Completion, even in Zero- or Few-Shot paradigms. However, they are known to hallucinate answers, or output results in a non-deterministic manner, thus leading to wrongly reasoned responses, even if they satisfy the user’s demands. To highlight opportunities and challenges in knowledge graphs-related tasks, we experiment with three distinguished LLMs, namely Mixtral-8x7b-Instruct-v0.1, GPT-3.5-Turbo-0125 and GPT-4o, on Knowledge Graph Completion for static knowledge graphs, using prompts constructed following the TELeR taxonomy, in Zero- and One-Shot contexts, on a Task-Oriented Dialogue system use case. When evaluated using both strict and flexible metrics measurement manners, our results show that LLMs could be fit for such a task if prompts encapsulate sufficient information and relevant examples.
\keywords{Large Language Models \and Knowledge Graph \and Knowledge Graph Completion \and Task-Oriented Dialogue System \and Prompt engineering.}
\end{abstract}
\section{Introduction}

Knowledge Graphs (KGs) are defined as graphs of data intended to accumulate and convey knowledge of the real world~\cite{hogan2021}. Their nodes represent entities of interest and edges represent potentially different relations between these entities.
KGs are integrated into various systems to enhance their abilities of storing and processing information.

Task-Oriented Dialogue (TOD) systems, alongside chatbots, are conversational agents possessing capabilities of engaging in natural language dialogues with human users. Different from chatbots, TOD systems aim to solve the user’s specific tasks within certain domains~\cite{chen2017}.
In our previous work \cite{iga2023}, we focused on developing an ontology-enhanced TOD system equipped with a static KG capable of mapping the context of the discussion and storing relevant information. 
Numerous benefits stem from adopting this approach, including enabling concurrent threads of conversation within a single discourse and utilizing the KG to validate data as a proxy. Additionally, the system gains the capability to execute Create-Retrieve-Update-Delete (CRUD) operations on domain-specific KGs. 
The acronym CRUD refers to the four basic operations that can be executed against persistent storages, such as relational or object databases, or other types of knowledge bases like KGs, to create, maintain or update them. As TOD systems aim to solve a variety of tasks, we decided that enabling such basic, but important ones is a suitable use-case to start with.
Specifically, we employed the Knowledge Graph Completion (KGC) task to construct two KGs and the Knowledge Graph Reasoning (KGR) task to handle CRUD operations. KGC’s objective is to deduce absent information within a specified KG~\cite{pan2023}, drawing from either input text or pre-existing knowledge.

Nonetheless, our TOD system relies on input text template-matching rules, constraining the authenticity of dialogues and hindering adaptability to novel concepts beyond the predefined ontology. 
Hence, in a subsequent study~\cite{iga2023-1}, we experimented with training neural networks - specifically fine-tuning BERT, a pre-trained model, to discern user intent and pertinent associated entities from input text. 
While embedding deep learning models into the TOD system architecture demonstrated encouraging outcomes, we still failed to completely address the previously mentioned limitations.

Therefore, in our current work, we study the use of LLMs to solve the KGC task, in the context of a TOD system.
 Literature~\cite{han2023,pan2023,zhang2021,zhu2023} identified a potential for synergy between KGs and LLMs, as KGs can enrich LLMs by supplying external knowledge for inference and explainability, while LLMs, in turn, can address KG-related tasks through natural language prompts.

Our experiments explore LLMs for static KGs contexts. Two well-established LLMs are used: Mixtral-8x7b-instruct-v0.1\footnote{\url{https://huggingface.co/mistralai/Mixtral-8x7b-Instruct-v0.1}}~\cite{mixtral}, and GPT-3.5-Turbo-0125\footnote{\url{https://platform.openai.com/docs/models/gpt-3-5-turbo}}, alongside the newest GPT model, the GPT-4o\footnote{\url{https://platform.openai.com/docs/models/gpt-4o}} version,  each possessing different properties. Communicating with such models involves the use of prompts, which are natural language instructions formatted in such way that the model understands the user’s intent. We test their capabilities of solving the aforementioned task using multiple prompting styles, including human-created and model-specific rephrased ones. Each prompt belongs to a level defined according to the TELeR taxonomy~\cite{teler2023}, that includes techniques such as Direct Prompting (DP), In-Context Learning (ICL), or Chain of Thought (COT), under Zero- and One-Shot contexts. To illustrate an appropriate application scenario for LLMs and KG tasks, we extract sample phrases from the training phase of our TOD system. Two datasets are obtained, including one with an increased difficulty, with test cases requiring reasoning steps that are not explicitly mentioned in the prompts. This approach allows us to not only evaluate the capability of LLMs in addressing the KG-specific tasks, but also to investigate their synergy with TOD systems. Finally, we report the accuracy and F1 scores of each LLM on both dataset, under strict and flexible measurement paradigms. 

Our research makes the following contributions. (i)	We assess the performance of two prominent LLMs: one open-source and the other proprietary, for the KGC task. This evaluation involves employing various prompts, either defined by humans or rephrased by the LLMs themselves, across different levels of complexity. We utilize three distinct prompting techniques (DP, ICL, COT) within two data contexts (Zero-Shot and One-Shot), yielding valuable insights into the capabilities of a robust LLM in performing such task. Metrics are measured within both strict and flexible paradigms, shedding light on the challenges encountered during post-processing. (ii) We introduce two personalized datasets tailored to gauge the performance of LLMs in the KGC task, featuring varying levels of difficulty. (iii) We investigate the feasibility of integrating such models into a domain-specific ontology-enhanced TOD system, by extracting and using test phrases specific to its context.


The paper evolves as follows. Section~\ref{seq:related} describes the related work about solving the KGC task  with LLMs. Section~\ref{seq:methodology} presents our methodology, describing the ingredients of our experiments. Section~\ref{seq:discussion} presents and discusses the results, while section~\ref{seq:conclusion} wraps up the paper with concluding remarks.

\section{Related Work}
\label{seq:related}

KGC aims to deduce fresh insights from existing knowledge within a KG or textual inputs. Ji et al.~\cite{ji2022} offer solutions for the KGC task utilizing embedding-based models like TransE, relation path reasoning exemplified by the Path-Ranking Algorithm, reinforcement-learning path finding, rule-based reasoning such as KALE, and meta relational learning utilizing R-GCN or LSTM. Similar insights are shared by Zhang et al.~\cite{zhang2021}, categorizing them into neural, symbolic, and neural-symbolic approaches. 

The aforementioned studies emphasize the usage of neural networks, logic networks, logic rules, or mathematical operations to address KGC. Interestingly, none of these endeavors particularly delve into the utilization of LLMs. Pan et al.~\cite{pan2023} explore the interplay between LLMs and KGs, proposing a unified framework that encompasses KCG too. Zhu et al.~\cite{zhu2023} experiment with ChatGPT and GPT4 for KGC, determining that while they lag behind state-of-the-art fine-tuned Pre-Trained Language models (PLMs) in a zero/one shot paradigm for completion, their reasoning capabilities often match or surpass those of SOTA models. Nevertheless, the comparative efficiency of an LLM versus a specialized PLM remains ambiguous. Han et al.~\cite{han2023} introduce PiVE, a prompting technique where a ChatGPT-based LLM extracts facts from input texts, while a smaller fine-tuned PLM iteratively verifies and supplements its responses. Wei et al.~\cite{wei2023} advocate for a multi-stage dialogue with ChatGPT to extract pertinent information from input texts, based on a predefined schema. 
Khorashadizadeh et al.~\cite{khorashadizadeh2023} explore the capabilities of foundation models such as ChatGPT to generate KGs from the knowledge it captured during pre-training as well as the new text provided to it in the prompt, grounded by several research questions. Their results show promising use cases for such models. 

As opposed to the above mentioned literature, we increase the number of textual inputs, expanding the generality of our conclusions.
Similar to Khorashadizadeh et al.~\cite{khorashadizadeh2023}, we test the capacity of  a proprietary LLM - namely GPT, with two versions: GPT-3.5-Turbo-0125 and GPT-4o on the KGC task. Moreover, we include an open-source LLM - Mixtral-8x7b-Instruct-v0.1~\cite{mixtral}, to facilitate research on open-source models, given their greater adaptability and cost-effectiveness compared to proprietary alternatives. To the best of our knowledge, we are among the first to test Mixtral for KG-related tasks. Another difference from them is that our prompts are more diverse and easier to track, as they are leveled according to the TELeR taxonomy~\cite{teler2023}. We introduce flexible metrics for gauging additional post-processing efforts. Finally, we also test the possibility of integrating an LLM with an ontology-enhanced TOD system, to sharpen its natural language processing and KG-related capabilities, by utilizing sample phrases from its training routine, resulting in two datasets, differentiated by their level of difficulty.

\section{Methodology}
\label{seq:methodology}

This section introduces our methodology used thorough this paper.
We describe the ontology used to anchor the LLMs knowledge, the datasets format and distribution, the prompt engineering steps, and the metrics measurement paradigms.

\subsection{Datasets format and distribution}

\begin{figure}[!bt]%
\centering
\includegraphics[width=0.96\columnwidth]{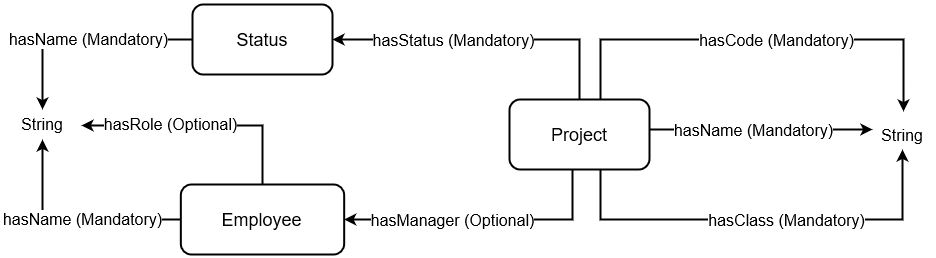}%
\caption{The ontology used throughout the experiments with three classes and six relationships}%
\label{fig:ontology}%
\end{figure}

Fig.~\ref{fig:ontology} depicts the ontology introduced in our prior research~\cite{iga2023} and used here. It comprises three classes: \textit{Project}, \textit{Employee}, and \textit{Status}, along with six relationships connecting them - such as \textit{hasManager} and \textit{hasStatus} or associating classes with literal values - like \textit{hasName}, \textit{hasRole}, \textit{hasClass}, and \textit{hasCode}. The ontology is described in RDF, using the Turtle syntax.

Input phrases are sourced from the training schedule of the TOD system developed in \cite{iga2023}, aimed to solve business operations around the concepts described in the above-presented ontology. Each phrase corresponds to the \textit{Create}  (\textit{Insert}) intent within CRUD operations, being focused either to one of the three available classes in the ontology or to other out-of-distribution (OOD) classes. Only few phrases resemble basic tasks without the \textit{Insert} intent. These, along with the OOD phrases, do not contain extractable triples and are labeled as having \textit{None} class type. The texts convey a various range of information, including both (i)~explicit information where intent, class type, associated relationships, and values are clearly articulated, and/or (ii)~implicit information where additional reasoning steps are required to identify the necessary details. For instance, texts may already provide an ID for the related instance, or the value might imply a name, role, or unspecified property. We also constructed phrases with misleading alternatives, containing grammatical errors or OOD class types. 

Table~\ref{tbl:examples} presents examples from each category. Details about the explicit types are available in the project’s repository\footnote{\url{https://github.com/IonutIga/LLMs-for-KGC}}.
Each text is accompanied by its related relationships, values, intent, and text type. Using regex templates, the information is converted into a dictionary. At the end, we obtained two datasets: \emph{Templates Easy} (TE) and \emph{Templates Hard} (TH). The first dataset  includes explicit and misleading text types, while the second one benefits from the addition of implicit-type  texts. 
Table~\ref{tbl:datasets} presents the distribution of texts per class type, on each dataset.

\begin{table}[!tb]
\caption{Input phrase examples and their types}
\scriptsize
\begin{tabular}{p{8.8cm}p{0.3cm}l}
\hline \hline
\textbf{Input phrase example} & & \textbf{Input phrase type} \\
\hline \hline
I want you to insert a project instance with code being A9I, its class is C, named BestApp and put Robert as the manager. & &	explicit information \\ \hline
Please put a project called UBBDemo identified by ZK5 managed by someone with something Mara and put it with other Python projects. & &	implicit information \\ \hline
Add a porject with code DS2, nme as Taskmate, class is Python and someone with role assistant as maager. & &	misleading information \\ \hline
I want you to insert an program instance with code being something like 0-Q7 its class is BASIC named UBBDemo and put Oscar as the manager. & &	misleading information \\
\hline \hline
\end{tabular}
\label{tbl:examples}
\end{table}

\begin{table}[!tb]
\centering
\caption{Datasets distribution of texts per class type - number of phrases}
\begin{tabular}{cccccc}
\hline
\multirow{2}{*}{Datasets} & \multicolumn{4}{c}{Class type}     & \multirow{2}{*}{Total} \\
        & Project & Employee & Status & None & \\
\hline
Templates Easy (TE)	&   58 & 	4	 & 3  &	7  &	72 \\
Templates Hard (TH) & 	56 & 	4	 & 3	& 15 &	78 \\
\hline
\end{tabular}
\label{tbl:datasets}
\end{table}

We associate each text with a set of golden labels, which are the target triples that can be extracted from the input text. Additionally, we consider that under the flexible metrics measurements paradigm (introduced in subsection~\ref{seq:metrics}), we can accept some triples as alternatives for  the golden ones,  i.e. some facts reported as false positives could be accepted if no other background information is available. It is widely acknowledged that extracting triples from text could yield a variety of  results depending on the expertise of the annotator. Therefore, we should allow room for LLMs to exhibit such variability.

To better illustrate the above description of the dataset, Fig.~\ref{fig:dictionary} presents an example of the obtained dictionary. The proposed alternative triple substitutes for triples two, three, and four. The accepted false positive triple refers to the user’s role, that can be inferred from the \textit{hasManager} relationship.

\begin{figure}[!tb]
\scriptsize
\begin{tabular}{|p{12cm}|}
\hline
$\{$ 
 'text': 'i want you to insert a project instance with code being A9I its class \\
\quad\quad\quad\quad		is C named BestApp and put Robert as the manager.', \\
 'golden$\_$labels': "$[\{$'subject': 'Project1', 'relationship': 'rdf:type', 'object': 'Project'$\}$, \\
\qquad\qquad\qquad  $\{$'subject': 'Project1', 'relationship': 'hasManager', 'object': 'Employee1'$\}$, \\
\qquad\qquad\qquad  $\{$'subject': 'Employee1', 'relationship': 'rdf:type', 'object': 'Employee'$\}$, \\
\qquad\qquad\qquad  $\{$'subject': 'Employee1', 'relationship': 'hasName', 'object': 'Robert'$\}$, \\
\qquad\qquad\qquad	$\{$'subject': 'Project1', 'relationship': 'hasCode', 'object': 'A9I'$\}$, \\
\qquad\qquad\qquad  $\{$'subject': 'Project1', 'relationship': 'hasClass', 'object': 'C'$\}$, \\
\qquad\qquad\qquad  $\{$'subject': 'Project1', 'relationship': 'hasName', 'object': 'BestApp'$\}]$", \\
 'alternative$\_$labels': "$[\{$'subject': 'Project1', 'relationship': 'hasManager', 'object': 'Robert'$\}]$",\\
 'fp$\_$ok$\_$labels': "$[\{$'subject': 'Employee1', 'relationship': 'hasRole', 'object': 'Manager'$\}$, \\
 \qquad\qquad\qquad $\{$'subject': 'Employee1', 'relationship': 'hasRole', 'object': 'manager'$\}]$", \\
 'class$\_$type': 'Project',\\
 'text$\_$type': 'MPS'$\}$\\
\hline
\end{tabular}
\caption{Example of a dictionary object with its text-related details.}%
\label{fig:dictionary}%
\end{figure}


\subsection{Prompt engineering}

To better encapsulate information and facilitate the replication of experiments, we utilize regex templates to convert each dictionary object - representing a text from a dataset, into a Prompt object. Such objects hold many details, such as the system prompt, its version and level, the input text, its type and  mentioned class type, the golden labels and the ones that could be accepted alternatively, the system message order, and metadata information about each model’s prediction of the input prompt. For a more comprehensive analysis, we enable the adjustment of the system's message position within the final prompt. Practice suggests\footnote{For prompt engineering we also followed OpenAI (\url{https://community.openai.com}) and HuggingFace (\url{https://huggingface.co/docs/transformers/main/tasks/prompting}) suggestions} that positioning the system's prompt after the user’s message could potentially enhance the performance of LLMs by mitigating long-context memory limitations. Lastly, we provide the option to flatten each Prompt object into dictionaries that are placed in text files for future reuse.

Three important paradigms are usually employed when designing prompts~\cite{zhao2023}: Zero-, One- and Few-Shot. The former’s prompt includes only the objective’s description and the input data that should be processed. One-Shot includes exactly one example of how the task should be solved against some different input data. Intuitively, Few-Shot refers to multiple, relevant examples added to the prompt.
After selecting the paradigm, one should decide about the prompting technique~\cite{zhao2023}. The current work employes three main approaches, as follows: Direct Prompting which refers to a prompt that only comprises the task’s description and the input to work on; In-Context Learning (ICL) which adds a relevant example of a solution to the given task on a different input data, and Chain of Thought (COT) which expands the prompt with the exemplified solution’s intermediary reasoning steps.

To test the model’s capacity to solve a task, we follow the guidelines of~\cite{teler2023} by assigning a level to each version of a system prompt. Specifically, we utilize levels 1 through 4 as outlined in~\cite{teler2023}. Within the fourth level, we further divide it into 4.1 and 4.2 to accommodate both ICL and COT variations of the prompt.

For better understanding, we exemplify the first level in Fig.~\ref{fig:l1prompting}. It sets the model’s role  as a KG expert, followed by instructions regarding the provided ontology. Subsequently, we outline the task at hand along with formatting guidelines for each instance’s ID and triple. Finally, the required output pattern is presented, attaching the target ontology. Level 2 adds a directive about the addition of the \textit{rdf:type} relationship. It then evolves into level 3, where we append a detailed bullet list of sub-tasks to be performed. All these levels adhere to the Zero-Shot paradigm, while levels 4.1 and 4.2 emulate ICL and COT, respectively, in a One-Shot manner. Depending on the golden labels of the target input text, we include either an example with no output triples or one with existing golden labels to better suit the specific scenario.

Moreover, as suggested in~\cite{pan2023}, we ask each model to rephrase the existing system prompts to better suit their needs. Therefore, we end up with two types of prompts: hand-written, and model-rephrased.

\begin{figure}[!tb]
\scriptsize
\begin{tabular}{|p{12cm}|}
\hline
'You are a Knowledge Graph Expert. A domain ontology is provided to you, delimited by double quotes.
	 The syntax used to describe the ontology is Turtle. Your input is a natural language text.
	 The input text may or may not contain references to instances of classes provided in the ontology, together with specific relationships.
	 Given the provided ontology, your task is to extract triples about the mentioned instances from the input text.
	 Each instance should be identified by an ID, using the format "Class" + "1",
		where "Class" is the name of the detected class and + is concatenation.
	 Put each triple in a JSON object, as follows: $\{\{$"subject" : ID, "relationship" : value, "object" : value$\}\}$.
	 If any triple refers to another instance, add all triples you assumed of that instance too.
	 Respond only with the JSON object(s) in a list.
	If no triple is detected, output "None". \textbackslash n
	Provided ontology: $\{$ontology$\}$ \textbackslash n' \\
	\hline
\end{tabular}
\caption{The level 1 system prompt.}%
\label{fig:l1prompting}%
\end{figure}


\subsection{Metrics}
\label{seq:metrics}

To gain a deeper understanding of the models' performance, we measure their accuracy and F1 score on both datasets. We employ two paradigms, namely strict and flexible metrics measurements.
 
Under the strict criterion, metrics are calculated in a standard text extraction manner, specifically counting how many predicted triples are among the golden ones, adhering to identical formatting. This approach enables the assessment of a model's ability to exactly follow the given prompt and process the input text, such that its results can be directly used in subsequent pipelines. 

The flexible paradigm proposed here allows  formatting mistakes that can be corrected in post-processing steps, or triples that are partially true to be counted as being accurate. This flexible measurement allows one to positively evaluate models that might not be such precise, but require fewer resources than more elaborate ones. 

Our flexible measurement paradigm modifies the strict calculation of the accuracy and F1 scores by including certain penalty for each instance that could be considered valid, even if it does not exactly follow the given prompt.  
In cases of erroneous formatted output, deductions are computed from the accuracy (100\%) of the entire prediction for each input text, whereas inaccuracies in the triple content lead to deductions only concerning the accuracy of a certain fact.
Below, we list the penalties we considered for each sort of LLM output errors.

Starting with the output format, the prompt demands a reply comprising solely a list of triples adhering to the specified template. Therefore, we consider a penalty of $2.5\%$ for outputs with multiple lists and a penalty of $7.5\%$ for the LLM producing additional text. 
If the prompt asks not to include the full IRI of an entity i.e. without the namespace, we penalize each addition with $1\%$. Finally, if a triple is output but does not contain exactly the three necessary keys, a penalty of $10\%$ is considered.  

Another category of penalties is related to the information content of a triple. For example, let's ask a model to construct a simple ID for each given instance of a class - specifically, the capitalized name of the class concatenated with “1”. We have noticed that some models tend to replace the number “1” with another single digit. Thus, if altering the final digit of a predicted identifier to “1” signifies correctness for a triple, the model is subjected to a penalty of $33\%$.  This percentage  value adheres to the three-component structure of a fact, such that, if one part is wrong, while the other two are correct, the model should still gain benefit of its prediction. This method of evaluation only applies to the validity of a constructed ID(s). Any other type of mistake is not allowed, since it alters the factuality of the implied information.

The penalty values described above and considered in our paper fit our specific task. We emphasize that these values are not fixed, and someone who wants to adapt the flexible measurement paradigm for another task is free to change them, according with her specific needs.

As previously noted, we permit certain alternative triples to the designated correct ones to be regarded as valid. Specifically, in Fig.~\ref{fig:dictionary}, concerning the relationship labeled as \textit{hasManager} between a \textit{Project} instance and an \textit{Employee} instance, if a model predicts the value of the object to directly be the employee’s name, instead of creating an \textit{Employee} instance and assign its type and name, the substitution will be counted as being correct. Nevertheless, the flexible metrics will attribute only one-third of the replacement as being accurate, implicitly penalizing the model for deviating from the prescribed ontology and guidelines. Additionally, some false positive triples may be deemed true in the absence of background knowledge, such as inferring the role of an employee as being a manager from the \textit{hasManager} relationship, thus not counting them as being wrong during the calculation of the model’s precision.

\section{Results and discussion}
\label{seq:discussion}

In this section, we present the obtained results and discuss them in order to conclude about the paper research questions.

Experiments were conducted on Google Colab, utilizing a virtual machine equipped with two Intel Xeon CPU 2.20GHz processors.
We experimented with Mixtral-8x7b-Instruct-v0.1, GPT-3.5-Turbo-0125 and GPT-4o. Mixtral is open-source, leveraging the Mixture of Experts\cite{mixtral} architecture, consisting of eight sub-networks, each of 7B parameters, accounting for a total of 56B parameters. GPT-3.5-Turbo-0125 is a well-known proprietary model that represents a fine-tuned version of GPT 3, consisting of 175B parameters.
GPT-4o boasts over 200B parameters, being the latest OpenAI model, advertised as their best performer. 
For Mixtral-8x7b-Instruct-v0.1, we used the HuggingFace Serverless API endpoint, whereas for GPT-3.5-Turbo-0125 and GPT-4o queries were directed to OpenAI's official API.


Each experiment was iterated three times, involving aprox. 6750 prompts in total, with each run lasting approximately 120 minutes. Interaction with Mixtral consumed about $50\%$ of the experimentation time. 
GPT-4o generated a cost around 40USD, while the GPT-3.5-Turbo-0125 only about 5USD.
For Mixtral, the HuggingFace endpoint generated no cost. 
Each set of predictions could  be loaded, tested and visualized from the paper’s repository, available at \url{https://github.com/IonutIga/LLMs-for-KGC}.

We notice that for the GPT models, an extra post-processing step is required, after receiving the produced output. Due to their ability to generate JSON formatted output, it surrounds its response with a specific tag (i.e. "```json...```"). One solution is to include a guideline in the prompt to avoid this behavior, but very rarely, around $0.5\%$ of times, it still adds it. Thus, to ensure that prompts are identical for all models, and be sure that the tag is not present in the output,  we post-process the GPT output in our code. We do this to enable a fair analysis solely of the output text.

Tables \ref{tbl:TE-handwritten} to \ref{tbl:TH-modelrephrased} display the results per model and prompt level, considering both strict and flexible metrics measurement paradigms. The first two tables focus on the \textit{Templates Easy (TE)} dataset, while the latter ones on \textit{Templates Hard (TH)} dataset. Tables \ref{tbl:TE-handwritten} and \ref{tbl:TH-handwritten} display the results for the hand-written system prompts, while in tables \ref{tbl:TE-modelrephrased} and \ref{tbl:TH-modelrephrased}, each model had to rephrase the prompts beforehand. 

We highlight the most effective prompts types per model and level in bold. The overall best prompts per line are underscored, while the overall best prompts per level are printed in italics.
Several interesting conclusions are discusses below.

\begin{table}[!tb]
\centering
\caption{Results on Template Easy (TE) dataset, using hand-written system prompts}
\begin{tabular}{cc|c|c|c|c}
\hline \hline
\multicolumn{2}{c|}{Model}  				& Mixtral & GPT-3.5-Turbo & GPT-4o & Total \\
\hline
Level & Metric                      & strict | flexible          & strict | flexible  & strict | flexible & strict | flexible      \\
\hline \hline
\multirow{2}{*}{1}     & Accuracy &	0.23 | 0.47 &	0.34 | 0.45 &	0.61 | 0.63 & 0.39 | 0.52 \\
      & F1	     &  0.25 | 0.58	& 0.41 | 0.58	& 0.73 | 0.75 & 0.46 | 0.64 \\
			\hline
\multirow{2}{*}{2}			& Accuracy &	0.19 | 0.49	& 0.41 | 0.51	& 0.63 | 0.85 & 0.41 | 0.62 \\
		  & F1	     &  0.18 | 0.53	& 0.47 | 0.61	& 0.64 | 0.89 & 0.43 | 0.68 \\
			\hline
\multirow{2}{*}{3}			& Accuracy &  0.19 | 0.44	& 0.30 | 0.42	& 0.71 | 0.86 & 0.40 | 0.57 \\
		  & F1	     &  0.20 | 0.55	& 0.38 | 0.58	& 0.72 | 0.89 & 0.43 | 0.67 \\
			\hline
\multirow{2}{*}{4.1}   & Accuracy & \textbf{0.25} | 0.63	& \textbf{0.88} | \textbf{0.88}	&  \underline{\textbf{0.89}} | \underline{\textbf{0.91}} & \textit{\textbf{0.67}} | 0.81 \\
			   & F1       &	\textbf{0.25} | 0.63	& \textbf{0.88} | \textbf{0.88}	& \underline{\textbf{0.89}} | \underline{\textbf{0.91}} & \textit{\textbf{0.67}} | 0.81 \\
			\hline
\multirow{2}{*}{4.2}   & Accuracy &	0.19 | \textbf{0.69}	& 0.85 | 0.87	& \underline{\textbf{0.89}} | \underline{\textbf{0.91}} & 0.64 | \textit{\textbf{0.82}} \\
				 & F1       &	0.19 | \textbf{0.75}	& 0.85 | 0.87	& \underline{\textbf{0.89}} | \underline{\textbf{0.91}} & 0.64 | \textit{\textbf{0.84}} \\
\hline \hline
\multicolumn{2}{c|}{Total Accuracy}	& 0.21 | 0.54 & 	0.55 | 0.62	& 0.75 | 0.83 & 0.50 | 0.67 \\
\multicolumn{2}{c|}{Total F1}  	      & 0.22 | 0.61	& 0.60 | 0.72	& 0.78 | 0.87 & 0.53 | 0.73 \\
 \hline \hline
\end{tabular}
\label{tbl:TE-handwritten}
\end{table}

\begin{table}[!bt]
\centering
\caption{Results on Template Easy (TE) dataset, using model rephrased prompts}
\begin{tabular}{cc|c|c|c|c}
\hline \hline
\multicolumn{2}{c|}{Model}  				& Mixtral & GPT-3.5-Turbo & GPT-4o & Total \\
\hline
Level & Metric                      & strict | flexible          & strict | flexible  & strict | flexible & strict | flexible      \\
\hline \hline
\multirow{2}{*}{1}     & Accuracy &	0.38 | 0.50 &	0.39 | 0.46 &	\textbf{0.62} | 0.64 & 0.46 | 0.53 \\
      & F1	     &  0.42 | 0.59 &	0.47 | 0.57	& \textbf{0.73} | 0.75 & 0.54 | 0.64\\
			\hline
\multirow{2}{*}{2}			& Accuracy &	0.15 | 0.37	& 0.43 | 0.46 &	0.36 | 0.85 & 0.31 | 0.56 \\
		  & F1	     &  0.17 | 0.49	& 0.51 | 0.58 &	0.36 | 0.90 & 0.35 | 0.66 \\
			\hline
\multirow{2}{*}{3}			& Accuracy & 0.20 | 0.42& 	0.39 | 0.46 &	0.07 | 0.77 & 0.22 | 0.55 \\
		  & F1	     &  0.22 | 0.51	& 0.49 | 0.59 &	0.07 | 0.73 & 0.26 | 0.61\\
			\hline
\multirow{2}{*}{4.1}   & Accuracy & 0.19 | 0.58 &	\underline{\textbf{0.85}} | \underline{\textbf{0.89}} &	0.66 | \underline{\textbf{0.90}} &  \textit{\textbf{0.57}} | 0.79 \\
		   & F1       &	0.19 | 0.59	& \underline{\textbf{0.85}} | \underline{\textbf{0.89}}	& 0.66 | \underline{\textbf{0.88}} & \textit{\textbf{0.59}} | 0.79 \\
			\hline
\multirow{2}{*}{4.2}   & Accuracy &	\textbf{0.42} | \textbf{0.74}	& 0.84 | 0.89	& 0.31 | 0.87 & 0.52 | \textit{\textbf{0.83}} \\
											 & F1       &	\textbf{0.42} | \textbf{0.78}	& 0.84 | 0.89	& 0.31 | 0.87 & 0.52 | \textit{\textbf{0.85}} \\
\hline \hline
\multicolumn{2}{c|}{Total Accuracy}	& 0.27 | 0.52	& 0.58 | 0.63	& 0.40 | 0.81 & 0.42 | 0.65 \\
\multicolumn{2}{c|}{Total F1}  	      & 0.28 | 0.59	& 0.64 | 0.71	& 0.43 | 0.83 & 0.45 | 0.71 \\
 \hline \hline
\end{tabular}
\label{tbl:TE-modelrephrased}
\end{table}

\begin{table}[!tb]
\centering
\caption{Results on Template Hard (TH) dataset, using hand-written prompts}
\begin{tabular}{cc|c|c|c|c}
\hline \hline
\multicolumn{2}{c|}{Model}  				& Mixtral& GPT-3.5-Turbo & GPT-4o & Total \\
\hline
Level & Metric                      & strict | flexible  & strict | flexible  & strict | flexible & strict | flexible       \\
\hline \hline
\multirow{2}{*}{1}     & Accuracy &	\textbf{0.25} | 0.41	& 0.34 | 0.42	& 0.53 | 0.54 & 0.37 | 0.46\\
      & F1	     &  \textbf{0.28} | 0.48	& 0.41 | 0.53   &	0.62 | 0.64 & 0.44 | 0.55 \\
			\hline
\multirow{2}{*}{2}			& Accuracy &	0.08 | 0.30	& 0.43 | 0.50 &	0.54 | 0.72 & 0.35 | 0.51 \\
		  & F1	     &  0.08 | 0.35 &	0.49 | 0.59 &	0.55 | 0.75 & 0.37 | 0.56\\
			\hline
\multirow{2}{*}{3}			& Accuracy & 0.09 | 0.35 &	0.28 | 0.40	& 0.59 | 0.74 & 0.32 | 0.50 \\
		  & F1	     &  0.10 | 0.46	& 0.35 | 0.52 &	0.60 | 0.75 & 0.35 | 0.58\\
			\hline
\multirow{2}{*}{4.1}   & Accuracy & 0.15 | 0.47 &	\underline{\textbf{0.77}} | \underline{\textbf{0.77}} &	0.71 | \textbf{0.76} & \textit{\textbf{0.54}} | \textit{\textbf{0.67}} \\
		& F1       &	0.15 | 0.49	& \underline{\textbf{0.77}} | \underline{\textbf{0.78}}	& 0.71 | \textbf{0.75} & \textit{\textbf{0.54}} | \textit{\textbf{0.69}} \\
			\hline
\multirow{2}{*}{4.2}   & Accuracy &	0.00 | \textbf{0.47}	& 0.75 | 0.76	& \textbf{0.73} | 0.74 & 0.49 | 0.66 \\
											 & F1       &	0.00 | \textbf{0.56}	& 0.75 | 0.76	& \textbf{0.73} | 0.75 & 0.49 | 0.69 \\
\hline \hline
\multicolumn{2}{c|}{Total Accuracy}	& 0.11 | 0.39	& 0.51| 0.57 &	0.62 | 0.70 & 0.42 | 0.56 \\
\multicolumn{2}{c|}{Total F1}  	      & 0.12 | 0.47	& 0.56 | 0.65	& 0.64 | 0.73 & 0.44 | 0.61 \\
 \hline \hline
\end{tabular}
\label{tbl:TH-handwritten}
\end{table}

\begin{table}[!bt]
\centering
\caption{Results on Template Hard (TH) dataset, using model rephrased prompts}
\begin{tabular}{cc|c|c|c|c}
\hline \hline
\multicolumn{2}{c|}{Model}  				& Mixtral & GPT-3.5-Turbo & GPT-4o & Total \\
\hline
Level & Metric                      & strict | flexible          & strict | flexible  & strict | flexible & strict | flexible       \\
\hline \hline
\multirow{2}{*}{1}     & Accuracy &	\textbf{0.33} | 0.43 &	0.32 | 0.41 & \textbf{0.51} | 0.54 & 0.39 | 0.46  \\
      & F1	     &  \textbf{0.37} | 0.50	& 0.38 | 0.52 &	 \textbf{0.60}  | 0.64 & 0.45 | 0.55\\
			\hline
\multirow{2}{*}{2}			& Accuracy &	0.14 | 0.35 &	0.39 | 0.46 &	0.28 | 0.72 & 0.27 | 0.51 \\
		  & F1	     &  0.15 | 0.45	& 0.45 | 0.56	& 0.28 | 0.77 & 0.29 | 0.59 \\
			\hline
\multirow{2}{*}{3}			& Accuracy & 0.12 | 0.39 &	0.38 | 0.45 &	0.09 | 0.66 & 0.20 | 0.50 \\
		  & F1	     &  0.13 | 0.46	& 0.47 | 0.55 &	0.09 | 0.65 & 0.23 | 0.55\\
			\hline
\multirow{2}{*}{4.1}   & Accuracy & 0.07 | 0.47	& \underline{\textbf{0.71}} | \underline{\textbf{0.75}}	& 0.55 | \underline{\textbf{0.76}} & \textit{\textbf{0.44}} | 0.66 \\
										   & F1       &	0.07 | 0.48	& \underline{\textbf{0.71}} | \underline{\textbf{0.75}}	&0.55 | \underline{\textbf{0.75}} & \textit{\textbf{0.44}} | 0.66 \\
			\hline
\multirow{2}{*}{4.2}   & Accuracy &	0.31 | \textbf{0.56}	& 0.70 | 0.74	& 0.18 | 0.73 & 0.40 | \textit{\textbf{0.68}} \\
											 & F1       &	0.31 | \textbf{0.61}	& 0.70 | 0.74	& 0.18 | 0.74 & 0.40 | \textit{\textbf{0.70}} \\
\hline \hline
\multicolumn{2}{c|}{Total Accuracy}	& 0.19 | 0.44 &	0.50 | 0.56 &	0.32 | 0.68 & 0.34 | 0.56 \\
\multicolumn{2}{c|}{Total F1}  	      & 0.20 | 0.50	& 0.55 | 0.63 &	0.34 | 0.71 & 0.36 | 0.61 \\
 \hline \hline
\end{tabular}
\label{tbl:TH-modelrephrased}
\end{table}

\textbf{Elaborate instructions without examples do not necessarily yield better results.} 
Upon analyzing both types of prompts across all levels, it appears that augmenting the prompt with more information without examples does not consistently enhance performance. Level 3 prompts, when evaluated rigorously, exhibit an average decline of around $8\%$ in accuracy and F1 score compared to levels 1 and 2. When evaluated using more flexible metrics, the discrepancy diminishes to almost zero. GPT-4o tends to increase its performance with each level when using hand-written prompts, while it dramatically decreases it for the model-rephrased ones. The other two models consistently lower their metrics at the third level, especially Mixtral-8x7b, which can be attributed to the inclusion of explanatory text, as it strives to replicate the input text.


\textbf{ICL and COT prompting techniques lead to best results.} All models scored their best results when prompted at levels 4.1 and 4.2, no matter the dataset or prompting template. 
Only GPT-4o had its best results for strict metrics at the first level when prompts where model-rephrased, which could be attributed to poor paraphrasing for the latter levels.
It is no surprise that such models work best when an adequate output example is given, as literature~\cite{teler2023} suggests. However, as Mixtral-8x7b sometimes provides explanations for its output, erroneous reasoning steps are noticeable, especially in cases where the input text contains a class type that is not present in the ontology. Thus, despite the GPT models exhibiting this behavior less frequently, LLMs still have significant room for improvement in terms of reasoning capabilities.

\textbf{Mixtral-8x7b rarely follows the required output format.}  The two metric measurement paradigms offer valuable insights into the models capacity to follow the given prompts. While GPT 3.5-turbo exhibits minimal disparity between the two perspectives, Mixtral 8x7b rarely produces texts that align with the specified template. Common errors include the addition of explanatory text (as evidenced by the 0 scores at the 4.2 level in table~\ref{tbl:TH-handwritten}) or the full IRI of an entity. When strictly evaluated, the open-source model only tops $42\%$ accuracy on the Template Easy (TE), while on flexible paradigm it reaches $74\%$ accuracy on the same dataset. GPT-3.5 outputs $88\%$ accuracy under both metrics measurements, while the GPT-4o variant yields $89\%$.

\textbf{Asking models to rephrase the system prompt might generally be a good idea for Mixtral-8x7b.} Some experiments in the literature~\cite{pan2023} ask the LLMs to formulate prompts for a given task. Inspired by it, we ask the LLMs to rephrase our manually written prompts to better align with their capabilities. As a comparison, Mixtral-8x7b benefits the most under rigorous evaluation, with an average increase of $14\%$ for each accuracy and F1 score. GPT-3.5-Turbo seems to conserve its behavior, signaling an increase of only $2.5\%$.
Surprisingly, GPT-4o exhibit a significant decrease in performance when it paraphrased the input prompts. On average, it lowered its performance by $33\%$ for both metrics, with third level prompts being the worst affected.
 Nonetheless, it's promising to see the open-source model enhancing its output by closely adhering to the provided system prompt.

\textbf{Implicit reasoning poses challenges for LLMs.} Template Hard (TH) dataset contains cases where the LLM needs to understand that a given value is already an ID that references an existing instance in a KG or that a statement implies a specific relationship pertaining to a class. As concluded by the results presented in the tables, under flexible metrics, Mixtral 8x7b achieves an accuracy of $56\%$ and an F1 score of $61\%$ on the more difficult dataset, which is $17.5\%$ lower than its performance on the easier one. GPT-3.5-Turbo narrows this margin, reducing from a peak accuracy and F1 score of $89\%$ to $78\%$ on Template Hard (TH). Same behaviour is observed with GPT-4o, as it falls from $91\%$ accuracy and F1 score to around $76\%$.
Interestingly enough, Mixtral-8x7b yields its best scores at level 1 prompts, when strictly measured. 

\textbf{GPT-4o is more consistent and performant, while GPT-3.5-Turbo achieves the best results on the harder dataset.}
Despite showing fluctuations in results when it rephrased the prompts, GPT-4o was the best overall model. On the TE dataset, under strict measurements, it had $75\%$ accuracy and $78\%$ F1 Score, almost four times more than Mixtral-8x7b and with $19\%$ more than GPT-3.5-Turbo. We can interpret the results as GPT-4o is more reliable than the other two models, regardless of the prompt level. However, GPT-3.5-Turbo came close to it considering their top performances, being only $3\%$ away from GPT-4o on the TE dataset, while surpassing it by $4\%$ on the TH dataset. Depending on the user's objectives, while considering the model's costs, the choice of the final model could vary.

In summary, KGC remains a challenging task for LLMs under Zero-Shot prompting. As models become better, they performance tend to increase, while shifting the focus on optimizing the costs.
Moreover, when checking their intermediate reasoning steps, they show lack of ability to follow the provided ontology. The open-source model has difficulties in conforming to the required output format. However, One-Shot contexts give promising results as LLMs excel in emulating a provided example. This implies that a less resource-intensive Few-Shot training approach could potentially boost performances, with a focus on techniques like Retrieval-Augmented-Generation to select more suitable examples within a given prompt. Another plus is their ability to enhance their inner knowledge to detect some implicit relationships from the input text. Nevertheless, as suggested by Fill et al.~\cite{fill2023}, presently we may use such LLMs as helpful assistants for solving such tasks, rather than ultimately faithful extractors in a pipelined system.

\section{Conclusion}
\label{seq:conclusion}

The proposed experiments showcases the ability of three leading LLMs, namely Mixtral-8x7b-Instruct-v0.1, GPT-3.5-Turbo-0125 and GPT-4o, in tackling the Knowledge Graph Completion task. 
Using both hand-written and model-rephrased prompts, we incorporated various prompt engineering techniques, such as In-Context Learning or Chain of Thought, focusing on Zero- and One-Shot contexts. Metrics measurement enabled the evaluation of the LLM strictly following the given prompt, as well as the its flexibility in considering post-processing steps. 
The results obtained from two distinct datasets tailored to various reasoning challenges highlight the LLMs strengths and weaknesses. These include their adaptability in Zero- or One-Shot scenarios and their utilization of internal knowledge to deduce implicit reasoning steps. However, they still lack self-awareness, not being able to adhere to explicit guidelines in the given prompt, or fully understand and exploit the considered ontology.

Additionally, we proposed two personalized datasets capable of assessing both the models' ability to solve the Knowledge Graph Completion task and their potential integration with task oriented dialogue systems simultaneously and a flexible measurement procedure to measure the capacity of the LLM to give logically correct results, but in an approximate format. 

Future work will prioritize the integration of additional LLMs for testing, facilitated by our interface's seamless incorporation of new endpoints. Moreover, we plan to test the possible influence of placing the system prompt at the end of the message, after the input text, to mitigate long-context memory issues. Lastly, we plan to move from single turns to a dialogue context, where the extraction happens as a discussion between a user and the LLM.

%
%
%
%
\bibliographystyle{splncs04}
\bibliography{references}

\end{document}